\pdfoutput=1

\documentclass[11pt]{article}

\usepackage{emnlp2021}

\usepackage{times}
\usepackage{amsmath,amssymb}
\usepackage{latexsym}
\usepackage{multirow}
\usepackage{graphicx}
\usepackage{breqn}
\usepackage[T1]{fontenc}

\usepackage[utf8]{inputenc}

\usepackage{microtype}

\title{Learning grounded word meaning representations on similarity graphs}

\author{Mariella Dimiccoli\\
Institut de Rob\`otica i Inform\`atica Industrial\\
  (CSIC-UPC), Barcelona, Spain \\
  {\small mdimiccoli@iri.upc.edu} \And
Herwig Wendt\\
  CNRS, IRIT\\
  Univ. of Toulouse, France \\
 {\small herwig.wendt@irit.fr}\\ \And
 Pau Batlle \thanks{\hspace{2mm}Work done during an internship at the IRI (CSIC-UPC).} \\
 California Institute of Technology\\
  Pasadena, California \\
 }

\begin{document}
\maketitle

\begin{abstract}
This paper introduces a novel approach to learn visually grounded meaning representations of words as low-dimensional node embeddings on an underlying graph hierarchy.
The lower level of the hierarchy models modality-specific word representations through dedicated but communicating graphs, while the higher level puts these representations together on a single graph to learn a representation jointly from both modalities. 
The topology of each graph models similarity relations among words, and is estimated jointly with the graph embedding.
The assumption underlying this model is that words sharing similar meaning correspond to communities in an underlying similarity graph in a low-dimensional space.
We named this model Hierarchical Multi-Modal Similarity Graph Embedding (HM-SGE).
Experimental results validate the ability of HM-SGE to simulate human similarity judgements and concept categorization, outperforming the state of the art. \footnote{\footnotesize Code available: \href{https://github.com/mdimiccoli/HM-SGE/}{https://github.com/mdimiccoli/HM-SGE/}.
Work partially funded by projects MINECO/ERDF RyC, PID2019-110977GA-I00, MDM-2016-0656.}
\end{abstract}

\section{Introduction}
During the last decade, there has been an increasing interest in deriving semantic representations from text corpora in terms of word vector space for both words in context and words in isolation \cite{mikolov2013distributed,ling2015finding,mccann2017learned,devlin2018bert,peters2018deep}. 
 However, semantic representations are tied to sensory experience, at least for concrete nouns \cite{anderson2017visually}, and deriving them by relying solely on text leads to lack of grounding on the extra-language modality. For this reason, several works have addressed the problem of grounding perceptual information in the form of visual information approximated by feature norms elicited from humans \cite{andrews2009integrating,silberer2012grounded}, or extracted automatically from images \cite{kiela2014learning,lazaridou2015combining,silberer2016visually}, or a combination of them \cite{roller2013multimodal}. 
Furthermore, several integration mechanisms for the linguistic and perceptual modalities have been proposed. They include 
methods employing transformation and dimension reduction on the concatenation of unimodal representations \cite{bruni2014multimodal,hill2014learning}; generative probabilistic models \cite{andrews2009integrating,roller2013multimodal,feng2010visual}; 
deep learning methods such as autoencoders and recursive neural networks \cite{silberer2016visually,socher2013zero}. 
While the approaches above take as input previously extracted unimodal representations, 
\cite{lazaridou2015combining,hill2014learning,zablocki2018learning} learn directly from raw data by integrating both modalities non-linearly in a low-dimensional space within a skip-gram model framework. These approaches typically lead to marginal improvements since only a small part of the vocabulary is covered by a corresponding visual information. Recent work has focused on generating pre-trainable generic representations for visual-linguistic tasks by using an instance-level contextualized approach   \cite{su2019vl,lu2019vilbert,sun2019videobert}. Such models have proved to be effective for several natural language applications such as visual question answering and visual commonsense reasoning, but their ability to simulate human behaviour phenomena 
has never been assessed so far. 


 In this paper, we propose a novel graph-based model for learning word representations across vision and language modalities that can simulate human behaviour in many NLP tasks.  
 Our assumption is that words sharing similar meaning correspond to communities in an underlying hierarchy of graphs in low-dimensional spaces. 
The lower level of the hierarchy models modality-specific word representations through modality-specific but communicating graphs, while the higher level puts these representations together  on  a  single  graph  to  learn  a  single representation jointly  from both  modalities.  At  each  level,  the  graph  topology models similarity relations among words, and it  is  estimated  jointly  with the graph  embedding. To the best of our knowledge this is the first model that uses a graph-based approach to learn grounded word meaning representations. Technically, our method is a joint feature learning and clustering approach. Moreover, it is compatible with both associative and domain-general learning theories in experimental psychology, following which the learning of a visual (linguistic) output is triggered and mediated by a linguistic (visual) input \cite{reijmers2007localization}, and these mediated representations are then further encoded in  higher-level cortices \cite{rogers2004structure}. This work has applications in cognitive science, prominently in simulations of human behavior involving deep dyslexia, semantic priming and similarity judgments, among others.


The contributions of this work are as follows: 1) We propose a novel technical approach to learn grounded semantic representations as low-dimensional embeddings on a hierarchy of similarity graphs, 2) we achieve state-of-the-art performance with respect to several baselines for simulating human behaviour in the tasks of similarity rating and categorization, 3) we demonstrate the ability of our model to perform inductive inference, 4) we validate the proposed approach through an extensive ablation study, and provide insights about the learnt representations through qualitative results and visualizations, 5) our model is compatible with associative and domain-general learning theories in experimental psychology.

\section{Related work}
Distributional Semantic Models (DSMs) are based on the distributional hypothesis \cite{harris2013papers}, following which words that appear in similar linguistic contexts are likely to have related meanings. DSMs associate each word with a vector (a.k.a.~word embedding) that encodes information about its co-occurrence with other words in corpora \cite{mikolov2013distributed,ling2015finding}. Recently, instance-level contextualized word embeddings \cite{devlin2018bert} have emerged as a natural extension of type-level non-contextualized DSMs and  have  demonstrated their effectiveness with respect to their counterpart in a wide variety of common NLP tasks \cite{mccann2017learned,devlin2018bert,peters2018deep}.


However, humans learn the verbal description of objects by hearing words while looking at /listening to/interacting with objects. 
Therefore, in recent years there has been an increasing interest in developing linguistic models augmented with perceptual information. These are commonly called \textit{grounded semantic spaces}. 
Following the classification introduced by \cite{collell2017imagined}, we can distinguish two integration strategies:
1) \textit{a posteriori combination}, where each modality is learnt separately and they are integrated afterwards
, and 2) \textit{simultaneous learning}, where a single representation is learnt from raw input data enriched with both modalities. 

\paragraph{A posteriori combination.}
Several works aimed at projecting directly vision and language into a common space. Among them, \cite{bruni2014multimodal} concatenate and project two independently constructed textual and visual spaces onto a lower-dimensional space using Singular Value Decomposition (SVD). Other approaches along the same line build on extensions of topic models as Latent Dirichlet Allocation (LDA), where topic distributions are learnt from the observed variables (words and other perceptual units) \cite{andrews2009integrating,roller2013multimodal,feng2010visual}. 
In \cite{kiela2014learning} an empirical improvement is obtained by using state-of-the-art convolutional neural networks to extract visual features, and the skip-gram model for textual features, that are simply concatenated. 
In \cite{silberer2016visually}  a stacked auto-encoder framework is used to  learn a representation by means of an unsupervised criterion (the minimization of the reconstruction error of the attribute-based representation) and then fine-tuned with a semi-supervised criterion (object classification of the input). In the approach proposed by \cite{wang2018learning} the weights of the unimodal feature concatenation are learnable parameters that allow to dynamically fuse representations from
different modalities according to different types of words. 
Other works focus on learning bimodal representations that are task-specific, 
with the goal of reasoning about one modality given the other \cite{lazaridou2014wampimuk,socher2013zero}. For example, the aim of image retrieval is to find a mapping between two modalities to tackle an image based task such as zero-shot learning \cite{frome2013devise} or caption generation/retrieval and caption-based image retrieval  \cite{socher2014grounded}. These models typically use a training criterion and an architecture suited to the task at hand.

\paragraph{Simultaneous learning.}
Little work has explored the possibility of learning multimodal representations directly from raw input data, i.e, images and corpora, building on the skip-gram framework. \cite{hill2014learning} treat perceptual input as a word linguistic context and has proved to be effective in propagating visual knowledge into abstract words.
\cite{lazaridou2015combining} modify the skip-gram objective function to predict both visual and linguistic features and is especially good in zero-shot image classification.  \cite{zablocki2018learning} contributed to this research line by leveraging the visual surroundings of objects  to fulfill the  distributional hypothesis for the visual modality.
This class of approaches typically leads only to a small empirical improvement of linguistic vectors since words from the raw text corpus associated
with images (and hence perceptual information) cover only a small portion of the training dataset.
In the last few years, increasing efforts are being devoted to deriving generic pre-trainable representations for visual-linguistic tasks based on transformers \cite{sun2019videobert,lu2019vilbert,su2019vl}.



\begin{figure}[t]
\small
\includegraphics[width=0.90\linewidth]{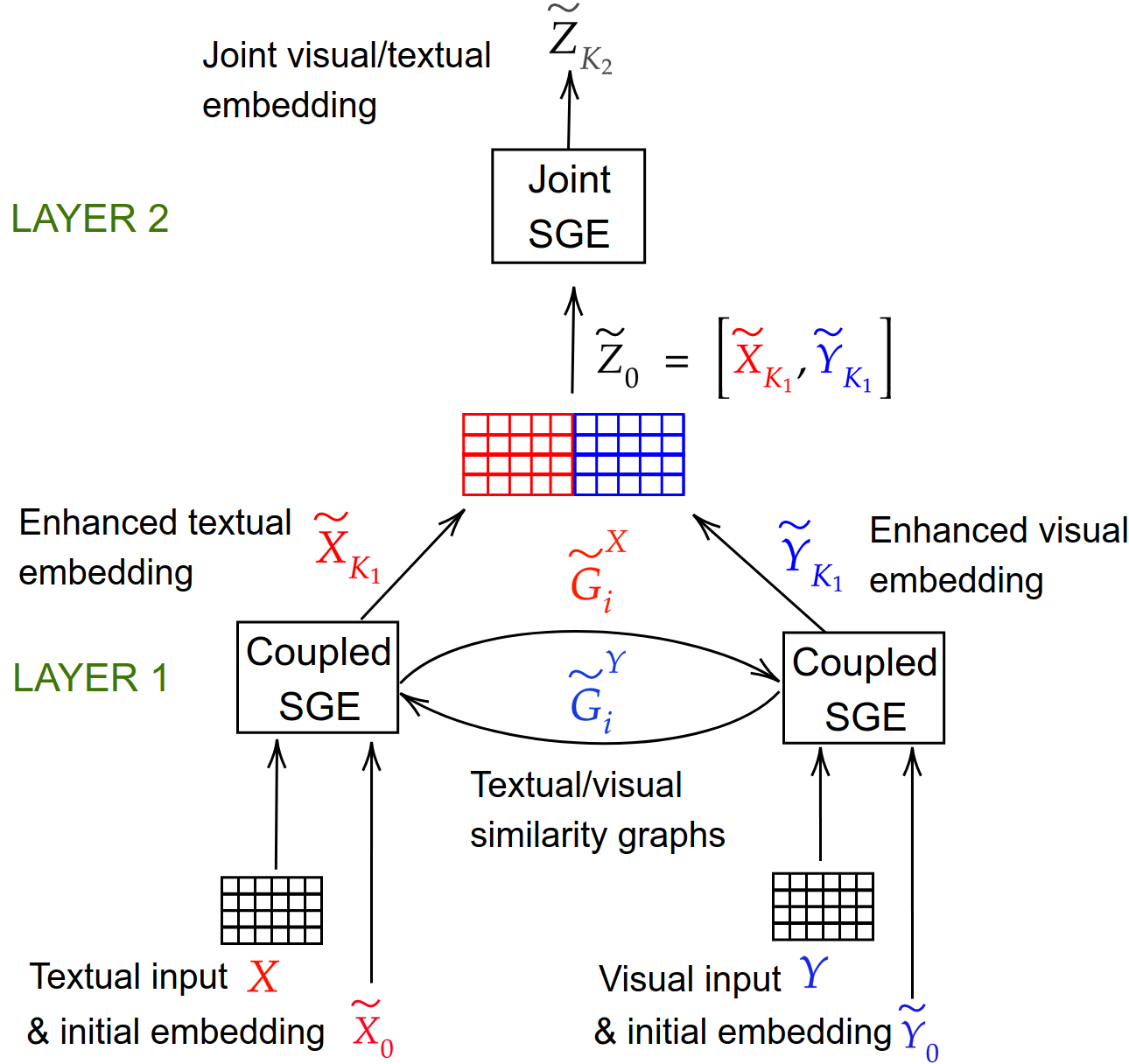}
\caption{HM-SGE model overview. The first layer learns unimodal representations conditioned to the other modality via its corresponding similarity graph. The concatenation of the enhanced representations is then used to learn jointly a common representation.}
\label{fig:overview}
\end{figure}

\begin{figure}[t]
\small
\includegraphics[width=0.90\linewidth]{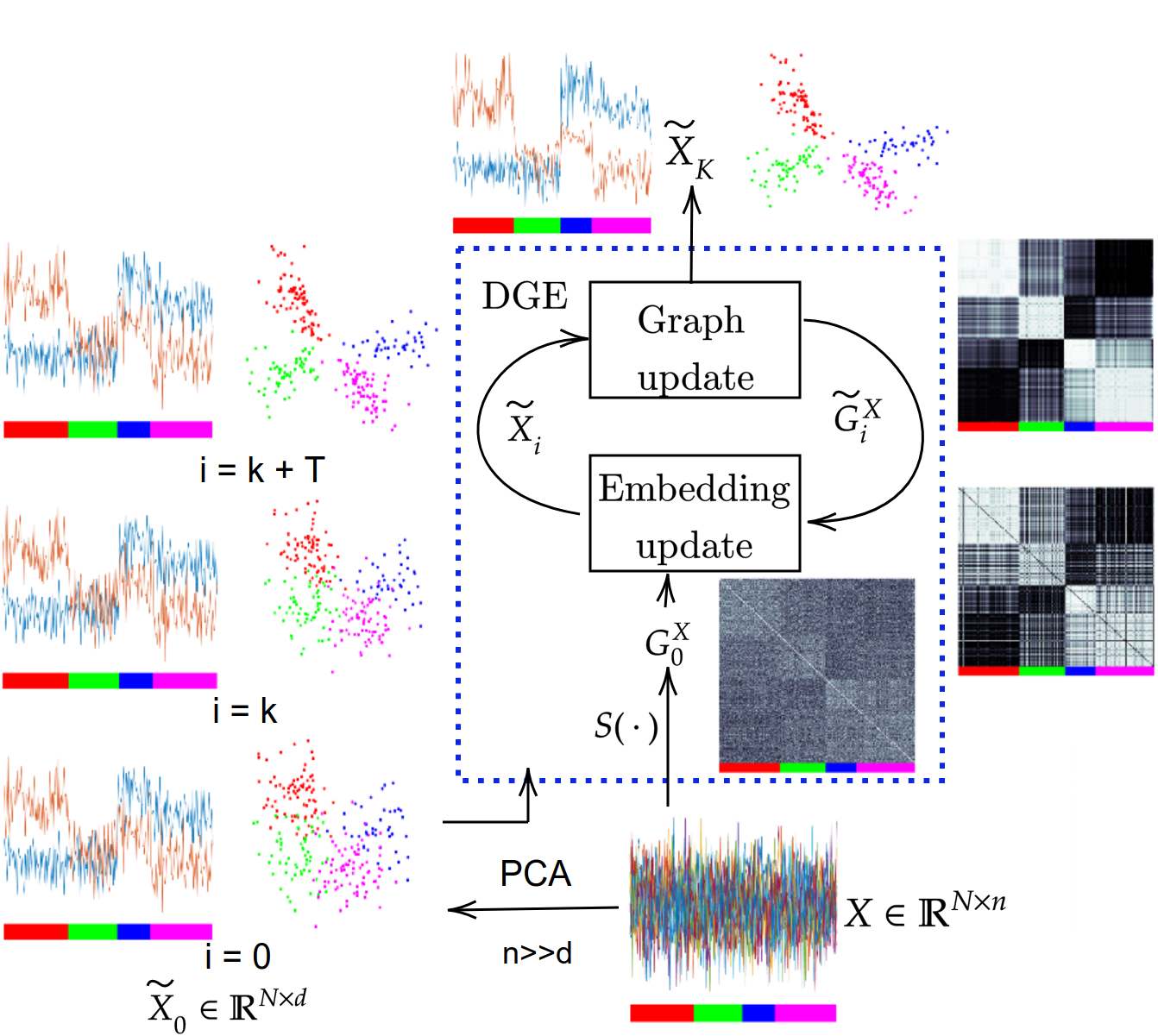}
\caption{DGE model of which our HM-SGE modules share the high-level structure. A similarity graph is estimated from the initial high dimensional embeddings that are then projected in a low-dimensional space. DGE alternates a graph embedding step with a graph update step.}
\label{fig:sge}
\end{figure}
\paragraph{Graph-based word meaning representations.}
Despite the success of graph-based models for sentence meaning \cite{koller2019graph},  their use for encoding word meaning representation has been little explored so far. Recently,  Graph Convolutional Network (GCN) based approaches have been used to improve text-based word meaning representation by incorporating syntactic information \cite{vashishth2018incorporating,tran2020syntactically,ryabinin2020embedding}. To the best of our knowledge, graph based models have never been exploited for grounded word meaning representations. 

\section{Proposed approach}

\paragraph{Model assumptions and psychological foundations.}
We assume that each modality, linguistic ($X$) and visual ($Y$), can be well represented by an unknown underlying graph in a low-dimensional space, in which nodes correspond to words and weighted edges to respective word similarity. 
The associative nature of human
memory, a widely accepted theory in experimental
psychology \cite{anderson2014human}, suggests that beyond being complementary, there exists a certain degree of correlation between these modalities \cite{reijmers2007localization}. 
Therefore, in a first step, our model aims at exploiting these correlations to enhance each unimodal representation individually.
This simulates the fact that when humans face, for instance, the task of visual similarity rating  of a pair of images, their mental search necessarily includes both components, the visual and the semantic one. 

In a second step, these enhanced unimodal representations are mapped into a common embedding space by inducing semantic representations that integrate both modalities.  This agrees with neuroscientific findings showing that it is unlikely that humans have separate representations for different aspects of word meaning \cite{rogers2004structure}.
Overall, the proposed two layer structure, where the first layer has two modality-specific branches that provide input to a second single-generic branch, is compatible with theories of both associative memory  \cite{reijmers2007localization} and  domain-general learning \cite{rogers2004structure}.




\noindent\includegraphics[width=\linewidth]{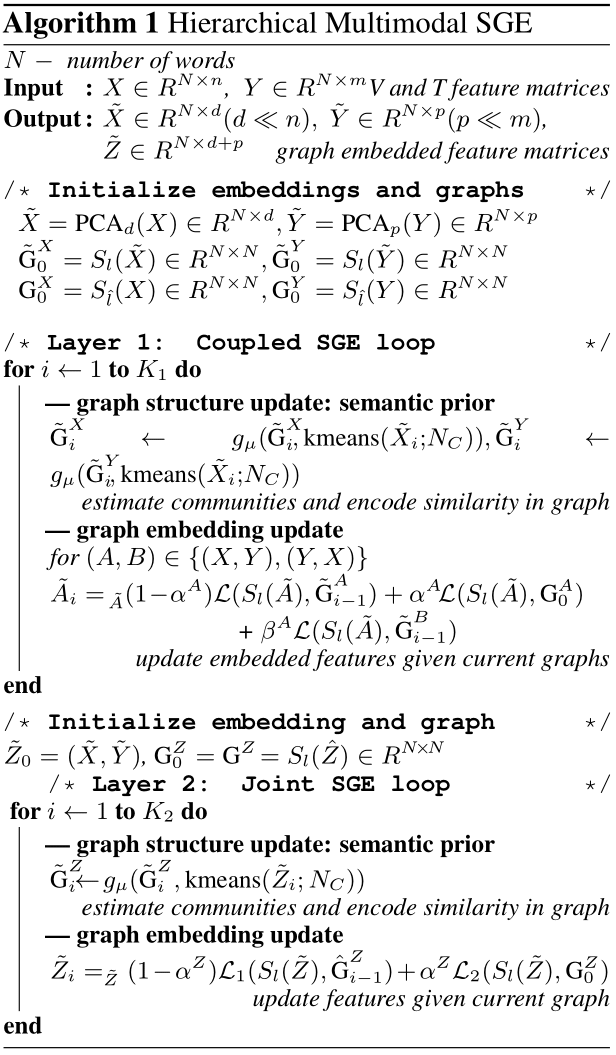}

\paragraph{\bf Architecture overview.}
Fig.~\ref{fig:overview} illustrates the proposed model. It takes as input linguistic ($X$) and visual ($Y$) vector representations.
 In the first layer, the initial embedding of both modalities are enhanced individually by relying on the other modality. In the last layer, the conditional embeddings are concatenated and jointly optimized.
For each modality, we first build a fully-connected similarity graph ${\textnormal{G}}^X_0$ from the initial features $X \in \mathbb{R}^{N\times n}$, where $N$ is the number of samples and $n$ is the feature dimension. ${\textnormal{G}}^X_0$ subsequently serves to regularize the process of jointly learning the underlying graph and an embedding $\tilde  X\in \mathbb{R}^{N\times d}$ of dimension $d\ll n$, which is achieved by 
alternating two steps:
at iteration $i$, 1) update node embeddings $\tilde  X_i$ by taking into account the current graph estimate $\tilde{\textnormal{G}}^X_{i-1}$ (reflected by edge weights) and that of the other modality, and  2) update the graph estimate $\tilde{\textnormal{G}}^X_{i}$ (fully connected similarity graph of $\tilde X_i$) by taking into account semantic similarity priors.

Conceptually, the alternating of a graph embedding step and a graph structure update step of our SGE is similar to the recently introduced Dynamic Graph Embedding (DGE) \cite{dimiccoli2020learning} (see Fig.\ref{fig:sge}), where a low-dimensional embedding is learnt from image sequences for the downstream task of temporal segmentation. However, we changed the way each of theses steps is formulated. Firstly, to learn unimodal (visual or textual) representations that take into account semantic communities in both graphs (textual and visual), we allow each modality to share its graph with the other modality. This is achieved by modifying the embedding loss during the embedding step. Secondly, since our data are not temporally linked, we do not model temporal constraints in the clustering step, but just semantic similarity among words.
Finally, we propose a two layers hierarchical architecture, that is tailored to the learning of visually grounded meaning representations.

\paragraph{Coupled similarity graph embedding update.}
Formally, the embedding update for $\tilde{X}$ (and analogously for $\tilde{Y}$) is computed as:
\begin{multline}
    \label{eq:embeddingHM}
\tilde{X}_i =   \arg\min_{\tilde{X}} (1\!-\!\alpha^X)\mathcal{L} (S_{ l}(\tilde{X}),\tilde{\textnormal{G}}^X_{i-1})\\
+\alpha^X\mathcal{L} (S_{ l}(\tilde{X}),\textnormal{G}_0) 
+ \beta^X \mathcal{L} (S_{ l}(\tilde{X}),\tilde{\textnormal{G}}^Y_{i-1}),
\end{multline}
where $\mathcal{L}$ is a cross-entropy loss function
that includes normalization of its arguments and $S_{l}$ stands for a cosine-distance based pairwise similarity function with exponential kernel of bandwidth $l$.
The first term in \eqref{eq:embeddingHM} controls the fit of the representation $\tilde X$ with the learnt graph $\tilde{\textnormal{G}}$ in low-dimensional embedding space, while the second term ensures that it keeps also aspects of the initial graph $\textnormal{G}_0^X$; $\alpha\in[0,1]$ controls the relative weight of the terms; the hyperparameters $(\beta^X, \beta^Y)$ tune the respective weights of the graphs of the other modalities in the unimodal representations.

\paragraph{Similarity graph update.}
To obtain an update for the graph at the $i$-th iteration, assuming that $\tilde{X}_{i}$ is given, the model starts from an initial estimate as $\tilde{\textnormal{G}}^X_{i} = S_{ l}(\tilde{X}_{i})$ and makes use of the model assumptions to modify $\tilde{\textnormal{G}}^X_{i}$. 
In particular, the semantic prior assumes that the most similar nodes form communities in the graph and leads to decreasing the edge weights between nodes of $\tilde{\textnormal{G}}^X_{i}$ that do not belong to the same community. Practically, this can be implemented by estimating communities using clustering, e.g. k-means with $N_C$ classes, and multiplying the edge weights for node pairs belonging to different communities by a factor $\mu\in(0,1)$; we denote this operation as
$\tilde{\textnormal{G}}^X_i\gets g_{\mu}\big(\tilde{\textnormal{G}}^X_i\!\!,\textnormal{kmeans}(\tilde X_i;N_C)\big)$.
%
%

%

\paragraph{Joint similarity graph embedding update.}
After $K_1$ iterations, the learnt representations are concatenated, $\tilde Z_0=(\tilde{X},\tilde{Y})$, and input to the second layer of our model. It learns the joint representation $\hat Z$ that integrates both modalities as node embeddings on an underlying graph encoding visually grounded word meaning. The last term of Eq. \eqref{eq:embeddingHM} is omitted at this stage. 
A detailed description of our framework is given in Algorithm 1.

\begin{figure*}[t]
\small\setlength{\tabcolsep}{0.1pt}
\centering\begin{tabular}{cccc}
tAttrib& vAttrib &  (tAttriv,vAttrib) & HM-SGE\\
\includegraphics[width=0.245\linewidth, trim=120 45 95 30, clip]{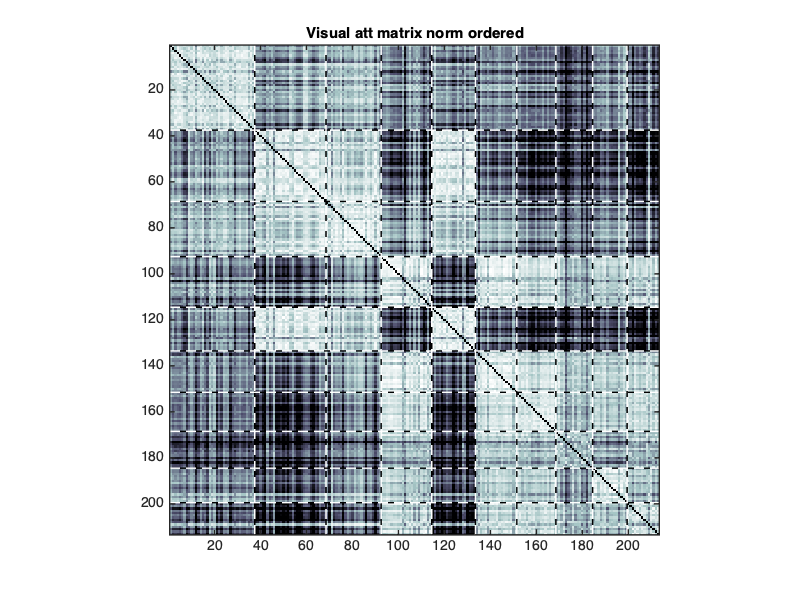}&
\includegraphics[width=0.245\linewidth, trim=120 45 95 30, clip]{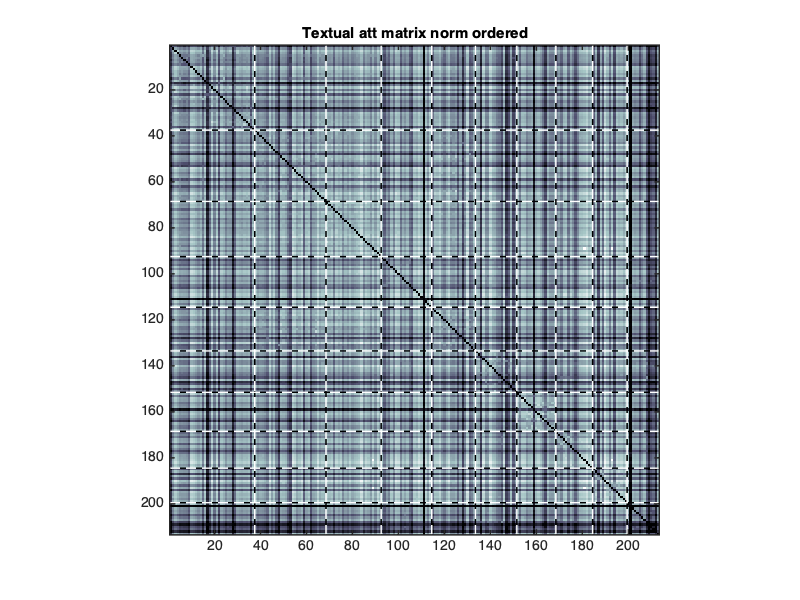}&
\includegraphics[width=0.245\linewidth, trim=120 45 95 30, clip]{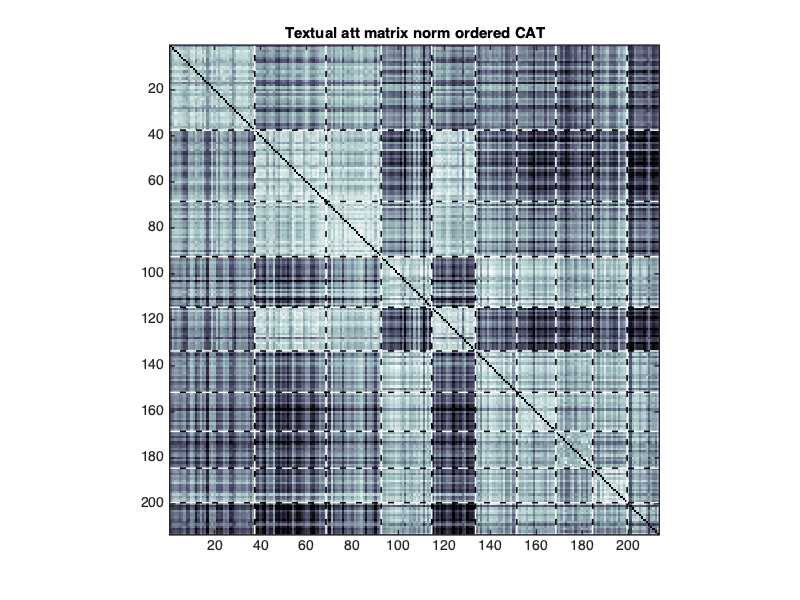}&
\includegraphics[width=0.245\linewidth, trim=120 45 95 30, clip]{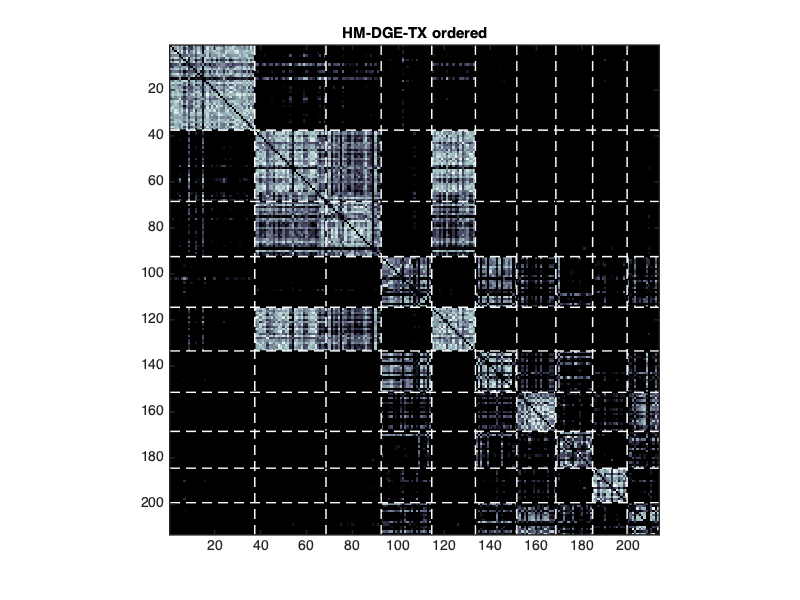}
\\
(a) & (b) & (c) & (d)\\
\end{tabular}
\caption{\label{fig:affinity}Similarity matrices of (a) textual attributes, (b) visual attributes, (c) visual and textual attributes (concatenation), (d) HM-SGE representations for the 10 categories {\it food, animal, bird, tools, mammal, weapon, instrument, transportation, clothing, device} (boundaries indicated by dashed lines). Image intensity values are individually clipped to cover the upper 3/8 of observed similarity values.}
%
%
\end{figure*}

\begin{table*}[t]
\small
\centering
\begin{tabular}{| l || c c c | c c c || c c c |}
\hline
Models &\multicolumn{3}{c|}{Semantic Similarity} &\multicolumn{3}{c||}{Visual Similarity} &\multicolumn{3}{c|}{Categorization}\\
& T & V & T+V & T & V & T+V & T & V & T+V \\ \hline
\hline
HM-SGE (tAttrib, vAttrib) &  \bf 0.74 & \bf 0.68 & \bf 0.77 & \bf 0.59 & \bf 0.60 &  0.64 & \bf 0.43 & \bf 0.39 & \bf 0.45   \\ 
\hline
SAE (tAttrib, vAttrib) & 0.67 & 0.61 & 0.72 & 0.55 & \bf 0.60 &  \bf 0.65 & 0.36 & 0.35 & 0.43 \\
GCN (tAttrib, vAttrib)     &  --- & --- & 0.74 & --- & --- &  0.59 &  --- &  --- &  0.42 \\ 
SVD (tAttrib, vAttrib) & --- & --- & 0.70 & --- & --- & 0.59 & --- & --- & 0.39 \\
CCA (tAttrib, vAttrib) & --- & --- & 0.58 & --- & --- & 0.56 & --- & --- & 0.37 \\
CONC (tAttrib, vAttrib) & 0.63 & 0.62 & 0.71 & 0.49 & 0.57 & 0.60 & 0.35 & 0.37 & 0.33 \\
CTL (tAttrib, vAttrib) & 0.67 & 0.64 & 0.69 & 0.54 & 0.57 & 0.58 & 0.38 & 0.29 & 0.30 \\
\hline
\hline
HM-SGE (skip-gram, vAttrib)  & \bf 0.79 & \bf 0.68 & \bf 0.78 & \bf 0.62 & \bf 0.60 &  0.65 & \bf 0.45 & \bf  0.40 &  0.47 \\ 
\hline
SAE (skip-gram, vAttrib) & 0.74 & 0.61 & 0.77 & 0.59 & \bf 0.60 &  \bf 0.66 & 0.44 & 0.35 &  \bf 0.48 \\
GCN (skip-gram, vAttrib)     &  --- &  --- &  0.76 &  --- &  --- &  0.60 &  --- &  --- &   0.42 \\ 
SVD (skip-gram, vAttrib) &--- & --- & 0.75 & --- & --- & 0.63 & --- & --- & 0.43 \\
CCA (skip-gram, vAttrib) &--- & --- & 0.59 & --- & --- & 0.57 & --- & --- & 0.35 \\
CONC (skip-gram, vAttrib) & 0.71 & 0.62 & 0.75 & 0.56 & 0.57 & 0.62 & 0.37 & 0.37 & 0.45 \\
CTL (skip-gram, vAttrib) & 0.74 & 0.67 & 0.71 & 0.60 & 0.57 & 0.60 & 0.44 & 0.29 & 0.43\\
\hline
\hline
VL-BERT &--- & --- & 0.66 & --- & --- & 0.56 & --- & --- & 0.36\\
Lazaridou et al. & 0.70 & 0.62 & 0.70 & 0.55 & 0.57 & 0.61 & 0.37 & 0.37 & 0.39 \\
Bruni et al. &--- & --- & 0.50 & --- & --- & 0.44 & --- & --- & 0.34 \\
\hline
%
\end{tabular}
\caption{\label{tab:comparison}Comparative results in terms of Spearman’s correlations between model predictions and human similarity ratings, and categorization on two sets of input features (tAttrib,vAttrib) and (skip-gram,vAttrib). Here T, V, T+V denote textual, visual and textual$\&$visual. The bold scores are the best results per semantic task.
Since the only stochastic part of our algorithm is the use of several random centroid seeds for kmeans, it can effectively be considered deterministic and results vary extremely little.
}
\end{table*}

\begin{table*}
\small
\centering
\begin{tabular}{| l || c c c | c c c || c c c |}
\hline
Models &\multicolumn{3}{c|}{Semantic Similarity} &\multicolumn{3}{c||}{Visual Similarity} &\multicolumn{3}{c|}{Categorization}\\
& T & V & T+V & T & V & T+V & T & V & T+V \\ \hline
\hline
HM-SGE (tAttrib, vAttrib)  &  --- & --- & \bf 0.77 & --- & --- & \bf 0.64 & --- & --- & \bf 0.45   \\ 
Layer2 only (tAttrib, vAttrib)   & --- & --- &  0.76 & --- & --- &  0.63 & --- & --- &  0.44 \\ 
Layer1 only (tAttrib, vAttrib)   & \bf 0.74 & \bf 0.68 &  0.76 & \bf 0.59 & \bf 0.60 &  0.63 & \bf 0.43 & \bf 0.39 &  0.44 \\ 
 tAttrib + vAttrib & 0.63 & 0.62 & 0.71 & 0.49 & 0.57 & 0.60 & 0.35 & 0.37 & 0.33 \\
\hline
HM-SGE (skip-gram, vAttrib)   & ---& --- & \bf 0.78 & --- & --- &  \bf 0.65 & --- & --- &  \bf 0.47 \\ 
Layer2 only (skip-gram, vAttrib) & --- & --- &  0.77 & --- & --- &  0.64 & --- & --- &   0.46 \\ 
Layer1 only (skip-gram, vAttrib) & \bf 0.79 & \bf 0.68 & \bf 0.78 & \bf 0.62 & \bf 0.60 &  0.64 & \bf 0.45 & \bf  0.40 &   0.46 \\ 
 skip-gram + vAttrib & 0.71 & 0.62 & 0.75 & 0.56 & 0.57 & 0.62 & 0.37 & 0.37 & 0.45 \\
\hline
%

\end{tabular}
\caption{Ablation study. T, V, T+V denote textual, visual and textual$\&$visual.  The best results are in bold.}
\label{tab:ablation}
\vspace{-2mm}
\end{table*}

\begin{table}[]
    \centering\small
\setlength{\tabcolsep}{1.4pt}
\begin{tabular}{|c|c|c|}
     \hline
       Word pairs 1-4 &Word pairs 5-8 & Word pairs 9-12\\
        \hline
        \hline
lettuce-spinach  & cello-violin &cloak-robe \\
clarinet-trombone &leopard-tiger & cabbage-spinach\\
cabbage-lettuce &raspberry-strawberry&pants-shirt\\
airplane-helicopter &chapel-church &blouse-dress\\
 \hline
    \end{tabular}
    \setlength{\tabcolsep}{3pt}
        \begin{tabular}{|l|c|}
     \hline
       Word clusters\\
        \hline
        \hline
       catfish, cod, crab, eel, guppy, mackerel, minnow, octopus\\
        perch, salmon, sardine, squid, trout, tuna\vspace{-0.18mm}\\
               \hline
       ambulance, bus, car, jeep, limousine, taxi, trailer, train\\
       truck, van\vspace{-0.18mm}\\
         \hline
                bike, buggy, cart, dunebuggy, motorcycle, scooter, tractor\\
                tricycle, unicycle, wagon\vspace{-0.18mm}\\
         \hline       
         ant, beetle, butterfly, caterpillar, cockroach, grasshopper\\
         hornet, housefly, moth, spider, wasp, worm\vspace{-0.18mm}\\
         \hline       
         apartment, barn, brick, bridge, building, bungalow, cabin\\
         cathedral, chapel, church, cottage, fence, gate, house, hut\\
         inn, pier, shack, shed, skyscraper\\
         \hline
    \end{tabular}
    \caption{Left: Word pairs with highest semantic and visual similarity according to HM-SGE
model. Pairs are ranked from highest to lowest similarity. Right: Examples of clusters produced by CW using semantic representations obtained with our HM-SGE.}
    \label{tab:similarity}
    \vspace{-2mm}
\end{table}

\setlength{\tabcolsep}{3pt}
\begin{table*}[]
\centering\small
\begin{tabular}{|l|l|l|l|}
     \hline
      \bf   Concept & \hfill Textual NN\hfill\textcolor{white}{.} &\hfill  Visual NN\hfill \textcolor{white}{.}&\hfill  HM-SGE NN\hfill\textcolor{white}{.} \\
        \hline
        \hline
            cabin        &   hut, tent & house, cottage, hut, bungalow  &  hut, cottage, house, shack, bungalow  \\
         \hline
           ox              &   cow & bull, pony, cow, calf, camel, pig, sheep, lamb & bull, cow, pony, sheep  \\
         \hline
           sardine              & tuna  & trout &  tuna, salmon, trout \\
         \hline
            bagpipe              &  accordion & clamp, accordion, tuba, faucet    &  accordion, tuba \\
               \hline
            hamster              & chipmunk, squirrel   &  rat, squirrel  &  squirrel, chipmunk, rat, groundhog \\
               \hline
            spoon              &  bowl  &  ladle, whistle, hammer   &  ladle  \\
               \hline
    \end{tabular}
    \caption{Nearest neighbors (NN, words with similarity larger than 0.9 times that of best pair), for textual, visual and MM-SGE vectors.}
    \label{tab:ablation:large}
\vspace{-2mm}
\end{table*}

\section{Experimental results}

\subsection{Experimental setting}

\paragraph{Visual and textual representations.}
As visual and textual input feature vectors we used the attribute based representations proposed by \cite{silberer2016visually}. Specifically, the visual modality is encoded via 414-dimensional vectors of attributes obtained automatically from images by training a SVM-based classifier for each attribute on the VISA dataset \cite{silberer2016visually}.
More specifically, we used initial meaning representations of words for the McRae nouns \cite{mcrae2005semantic} covered by the VISA dataset \cite{silberer2016visually}, that consists exclusively of concrete nouns.

The textual modality was encoded in two different ways: through textual attributes and via word embeddings.
Textual attributes were extracted by running Strudel \cite{baroni2010strudel} on the WaCkypedia corpus \cite{baroni2009wacky}, and by retaining only the ten attributes with highest log-likelihood ratio scores for each target word. The union of the selected attributes leads to 2,362 dimensional textual vectors.
Word embeddings were obtained by training the skip-gram model \cite{mikolov2013distributed} on the WaCkypedia corpus \cite{baroni2009wacky}, resulting in 500-dimensional embedding vectors.
The attribute-based representations in both modalities were scaled to the [$-1, 1$] range.

\paragraph{Hyperparameter settings.}
We use a common embedding dimension $d=p=15$ for both modalities. The SGE hyperparameters are set to $(\alpha^X,\mu^X,N_C^X)=(0.1, 0.95,25)$ and $(\alpha^Y,\mu^Y,N_C^Y)=(0.3, 0.7, 5)$ for the first layer of our model.
For the second layer of our model, we set $(\alpha^Z,\mu^Z,N_C^Z)=(0.05,0.7, 20)$ and $(0.1, 0.7, 6)$ when using textual or skip-gram as input, respectively. The number of SGE iterations are set to $K_1=4$ or $5$ (first layer) and $K_2=2$ or $5$ (second layer) when textual attributes or skip-gram representations are used for the textual modality, respectively. These hyperparameters have been optimized for each SGE individually using grid search; the cross-coupling parameters were also optimized separately and set to $(\beta^X,\beta^Y)=(0.01,0.1)$. 


%
%
%
%
%
%

\paragraph{Performance measures.}

Similarly to previous work \cite{silberer2016visually}, we evaluate our
model on two different semantic tasks, namely word similarity rating and categorization. 
Specifically, we measure how well our model predictions of word similarity correlate with human semantic and visual similarity ratings using Spearman’s correlation. Our similarity ratings are calculated as the cosine similarity of learnt representation vectors. As human similarity ratings, we used those published in \cite{silberer2016visually}.
The semantic categories are induced by following a clustering-based approach, namely the Chinese Whispers algorithm \cite{biemann2006chinese}, and the quality of the clusters produced  was evaluated using the F-score measure introduced in the SemEval 2007 task \cite{agirre2007semeval}.

\paragraph{Computation.}
The complexity of our approach is $\mathcal{O}(N^2)$. Experiments were conducted on a 2018 Dell Precision T7920 workstation with 64GB RAM and a single NVIDIA Titan XP GPU.

\subsection{Comparative results.}
In Tab.~\ref{tab:comparison}, we compare our HM-SGE to the state of the art.
Results are presented for two different sets of input features: attribute vectors for the visual modality (vAttrib) combined with either attribute vectors (tAttrib) or skip-gram encoding (skip-gram) for the textual modality (top and center part of Tab.~\ref{tab:comparison}, respectively). The bottom part of Tab.~\ref{tab:comparison} presents comparisons with models for raw data \cite{lazaridou2015combining} and \cite{bruni2014multimodal} and pre-trained VL-BERT model \cite{su2019vl} \footnote{https://github.com/jackroos/VL-BERT} for which we derived the type-level representations. Each column of Tab.~\ref{tab:comparison} corresponds to a unimodal (textual (T) or visual (V)) or joint (T+V) representation.

As baseline methods trained with the same attribute based input, we considered:
SAE \cite{silberer2016visually}, SVD \cite{bruni2014multimodal}, CCA \cite{hill2014learning} and cross-transfer learning (CTL) \cite{both2017cross} models trained on the same attribute-based input as in \cite{silberer2016visually}; SVD and CCA models first concatenate normalized textual and visual vectors and then conduct SVD or CCA; CONC stands for concatenation of normalized textual and visual vectors \cite{kiela2014learning}. CTL transfer information from the common space to unimodal representations by using either a mask (estimated by correlation, or multilinear regression) or a function generating artificial features (linear, or a neural net) estimated from multimodal examples. We implemented the mask approach and reported the better of the results for correlation or multilinear regression.

Moreover, we provide a novel graph-based baseline that learns word embeddings via a two-layer Graph Convolutional Networks (GCN) trained to classify words. The graph structure (edges) was created by using visual features, and node embeddings were initialized by textual features.

The unimodal (output by layer 1) representations of our HM-SGE always achieve state-of-the-art results, largely outperforming the SAE method (up to $+7\%$, $+4\%$ on average). 
The unified representations of our HM-SGE (output by layer 2 following layer 1) achieves state-of-the-art results in most cases:  for semantic similarity rating and categorization when using textual attributes, and for semantic similarity only when using the skip-gram model for the textual modality  (up to $+5\%$, $+3\%$ on average). In the other cases (visual similarity ratings; categorization for skip-gram model) the unified representations also achieve results comparable to the best performing method (SAE), up to 1$\%$ difference.  
Overall, we improved reported performance measures with respect to the SOTA, by up to 7$\%$ on average for the 18 cases, in particular by 5$\%$(semantic similarity), 1$\%$ (visual similarity) and 3$\%$ (categorization).

Our model also outperforms  embeddings obtained using the pretrained VL-BERT by a large extent. This is not surprising considering that recent studies \cite{mickus2021you,rogers2020primer} have raised concerns about the coherence of BERT (text-based) embedding space. Further, as one would expect, tAttrib dominates vAttrib when modeling semantic similarity and categorization, while vAttrib dominates tAttrib for visual similarity ratings only. More importantly, the joint use of tAttrib and vAttrib improves all evaluation metrics,  hence corroborating the fact that the model has learnt to leverage on their redundancy and complementarity. Joint representations also improve performance when based on skip-gram encoding, except for semantic similarity, which is strongly dominated by the skip-gram features.
Examples of our model output are given in Tab.~\ref{tab:similarity}, showing word pairs with highest similarity rating (left) and examples of word clusters (categories, right).


\subsection{Model validation and illustration}
\paragraph{Ablation study.}
To validate the proposed HM-SGE model, in  Tab.~\ref{tab:ablation} we report results obtained with HM-SGE (top rows), and with HM-SGE upon removal of one of its layers: removal of the two coupled SGE (second rows, Layer2), removal of final SGE (third rows, Layer1), no HM-SGE (bottom rows); for rows 2 to 4, joint representations (T+V) are obtained upon concatenation of the individual visual and textual representations. It can be seen that Layer1 as well as Layer2 alone yield significant performance improvements when compared with the initial representations (T, V and T+V). This validates the independent capabilities of the individual components of our model to learn meaningful word representations. Yet, best performance for the joint representation (T+V) are obtained only upon combination of the two layers, demonstrating the importance of both layers in our model. 

\paragraph{Qualitative results.}
An illustration corresponding to the rows 1 and 4 of Tab.~\ref{tab:ablation} is provided in Fig.~\ref{fig:affinity}, which plots the affinity matrices for textual attributes (T), visual attributes (V),  concatenation thereof (T+V) and HM-SGE for the 10 categories {\it food, animal, bird, tools, mammal, weapon, instrument, transportation, clothing, device}; category boundaries are indicated by dashed lines, and image intensity values are clipped to cover the upper 3/8 of observed similarity values for each representation individually. It is observed that T yields visually better results than V, and the concatenation T+V inherits similarity from both attributes but is dominated by T. The observed large off-diagonal similarity values for T, V and T+V lead to expect that categorization results based on these attributes are poor. The affinity obtained by HM-SGE is more structured, with large values  essentially coinciding with within-category affinities (the 10 diagonal blocks) and a few off-diagonal blocks. Interestingly, these off-diagonal blocks correspond with the arguably meaningful two groups of categories {\it animal, bird, mammal} (rows-columns 2,3,5) and {\it tools, weapon} (rows-columns 4,6). 

Finally, Tab.~\ref{tab:ablation:large} exemplifies such results and provides a different view by showing the nearest neighbors (NN) for six words in terms of similarity computed on textual attributes, visual attributes, and our HM-SGE attributes; all neighbors with similarity values of at least 90\% that of the closest neighbor are given. The examples illustrate that the unified word meaning representations learnt by HM-SGE lead to NN that are not a simple union or intersection of visual and textual NN, but that HM-SGE is capable of removing pairs that make less sense (e.g. {\it tent} for {\it cabin}; {\it calf,camel,pig} for {\it ox}; {\it clamp,faucet} for {\it bagpipe}; {\it bowl,whistle,hammer} for {\it spoon}) and can identify and add new meaningful pairs (e.g. {\it salmon} for {\it sardine}; {\it groundhog} for {\it hamster}). 

\paragraph{Inductive inference.}
It is possible to use our model to perform inductive inference when one of the two modalities for a concept, say modality $A$, is missing. To this end, it suffices to replace in Eq.~\eqref{eq:embeddingHM} the corresponding row and column of the matrix $\tilde{\textnormal{G}}_{i-1}^A$ with those of $\tilde{\textnormal{G}}_{i-1}^B$ for the other modality $B$. For example, when only the visual component $\tilde X$ for the concept {\it bluejay} is given, the textual representation $\tilde Y$ learnt by HM-SGE outputs the nearest neighbors {\it robin, stork, falcon, finch}, which are all birds; to give another example, from the visual attribute for {\it shelves} HM-SGE predicts textual nearest neighbors {\it dresser, cabinet, cupboard, bureau, desk, closet}. Analogously, HM-SGE predicts visual nearest neighbors {\it shelves, cabinet} and {\it dagger, knife, spear} for {\it cupboard} and {\it sword}, respectively, when only the textual attribute is given for these two concepts.

\section{Conclusion}
This paper has proposed a novel approach, named HM-SGE, to learn grounded word meaning representations as low-dimensional node embeddings on a hierarchy of graphs. The first layer of the hierarchy encodes unimodal representations conditioned to the other modality, and the second layer integrates these enhanced unimodal representations into a single one. The proposed HM-SGE approach is compatible with theories of associative memory \cite{reijmers2007localization} and of domain-general learning \cite{rogers2004structure}. 
Comparative results on word similarity simulation and word categorization show that our model outperforms baselines and related models trained on the same attribute-based input.
Our evaluation reveals that HM-SGE is particularly good at learning enhanced unimodal representations that simulate how the response of our brain to semantic tasks involving a single modality is always triggered by other modalities. Moreover, it succeeds in encoding these unimodal representations into a meaningful unified representation, compatible with the point of view of domain-general learning theory.
The ablation study thoughtfully validates the proposed hierarchical architecture. Beside quantitative results, we give several insights on the learnt grounded semantic space through visualization of nearest neighbors, clusters, and most similar pairs. These additional results corroborate the quality of the learnt multimodal representations.  Furthermore, the proposed approach is able to perform inductive inference for concepts for which only one modality is available.


\clearpage

\bibliography{anthology,custom}
\bibliographystyle{acl_natbib}

\end{document}